%% file: main-2.tex
\documentclass{article} 
\usepackage[preprint]{neurips_2026}
\usepackage{graphicx}
\usepackage{subcaption}
\usepackage{natbib}
\usepackage{booktabs}
\usepackage[ruled,vlined,linesnumbered]{algorithm2e}
\usepackage{bm}
\usepackage{amsmath}
\usepackage{url}

\author{%
 Alina Hyk \\
 Oregon State University 
 \And Sandhya Saisubramanian\\
 Oregon State University \\
}

\title{Learning Transferable Latent User Preferences for Human-Aligned Decision Making}

\begin{document}

\maketitle

\begin{abstract}
Large language models (LLMs) are increasingly used as reasoning modules in many applications, yet they often struggle to produce human-aligned decisions. Human-aligned decision-making requires accounting for both explicitly stated goals and latent user preferences that shape how
ambiguous situations should be resolved. Existing approaches either learn these preferences through continual per-interaction feedback, which is costly for both the user and the agent, or keep them implicit in dialogue transcripts or example sets tied to a single model. We present CLIPR (Conversational Learning for Inferring Preferences and Reasoning), a framework that represents user's preferences as an explicit, compact set of natural-language rules used at inference time for human-aligned decision-making.  CLIPR elicits these rules from conversational input, monitors decision accuracy, and consults the user when performance degrades unexpectedly. CLIPR revises the learned rules, based on the conversation, only if it does not reduce accuracy on scenarios already processed. Evaluations on three preference-ambiguous datasets show that CLIPR matches or outperforms baselines while using up to 96\% fewer LLM calls than continual-feedback preference learning. Our results also show that rules learned by one LLM transfer unchanged to others, and that our approach can recover from corrupted rules. In a user study with 30 participants, CLIPR outperformed baselines and was rated as requiring less effort than continual-feedback learning.

\end{abstract}

\section{Introduction}
Large language models (LLMs) are increasingly used as decision-making modules in human-facing systems, from virtual assistants to personal robots~\citep{han2025llmpersonalize,pmlr-v229-rana23a}. Their usefulness depends not only on understanding a user's explicit goal, but also on acting consistently with latent user preferences that shape how ambiguous situations should be resolved. Latent preferences are rarely specified in advance, making it difficult to produce decisions that are aligned with user preferences.

\begin{figure*}[t]
\centering 
\includegraphics[width=0.9\linewidth]{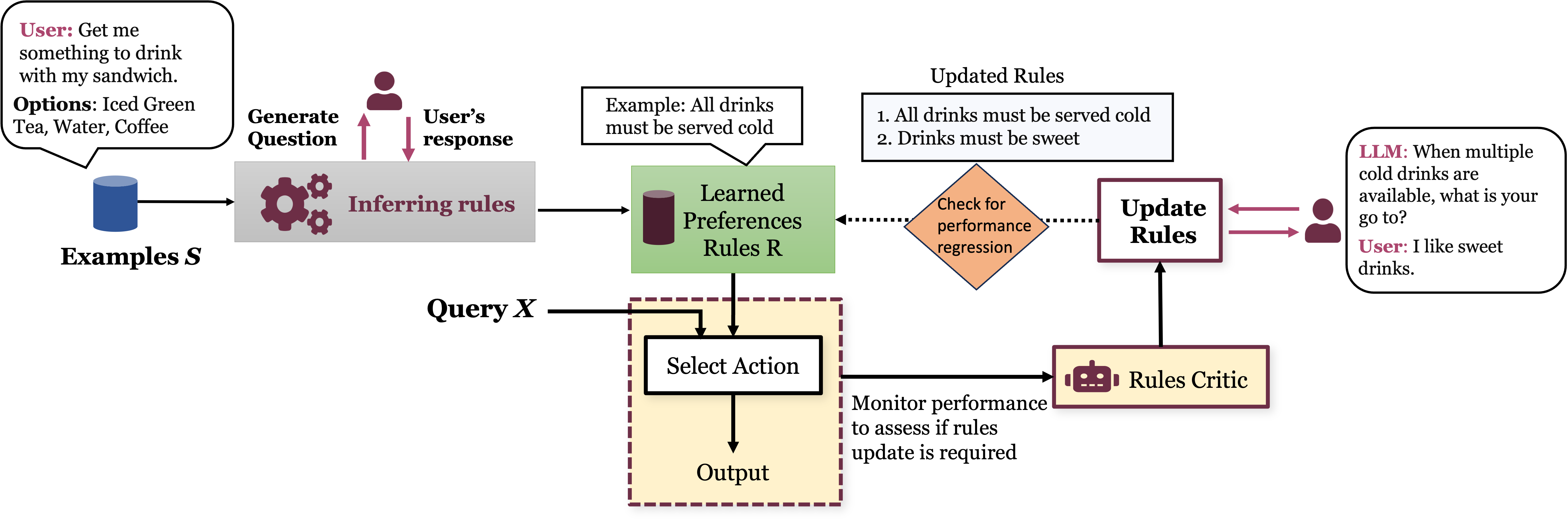}
\caption{Overview of CLIPR framework. The agent asks questions about preference dimensions that remain unresolved by the dialogue and example decisions, and compiles the resulting feedback into actionable rules that are used at inference time. The performance is monitored by a rules critic that determines when to seek additional information and the rules are revised based on the additional information, only if the past performance does not regress. }
\label{fig:Interactive_Rules}
\end{figure*}

This challenge is more pronounced in scenarios where multiple valid actions exist, but only one aligns with a user's preferences. Consider the request ``Get me something to drink with my sandwich'', which admits multiple valid responses: Iced green tea, water, and coffee (Figure~\ref{fig:Interactive_Rules}). Determining which one the user would prefer requires additional knowledge---for example, a latent preference for cold beverages. Such ambiguities are common, especially when users provide underspecified queries or lack experience in prompting LLMs~\citep{wang2024task}.  Without user preference information, selection among valid actions is effectively arbitrary, leading to misaligned outcomes.

Existing methods provide only partial solutions to this problem and broadly fall into three categories: (1) storing user interactions or information as retrievable memory or profiles (e.g., user.md-style systems)~\citep{zhou2025tom}; (2) repeatedly reasoning over interaction history and past corrections to infer preferences at inference time~\citep{gao2024cipher,li2025gate}; and (3) inferring preferences from non-conversational feedback types such as demonstrations, rankings, or examples~\citep{song2024preference,2023arXiv230405302Y,10.5555/3737916.3739580,wu2023tidybot}. These approaches can personalize behavior, but they either leave the agent's understanding of the user implicit or repeatedly incur the cost of reconstructing that understanding. Rewriting the preference model after each manual correction~\citep{tao2025lams} also makes the reactive adaptation brittle since a single ambiguous or misleading response can overwrite a preference model that was already working well. While in-context learning~\citep{dong2024survey} seems plausible as context window is expanding in recent LLMs, keeping the full history in context means the model must re-infer the user's preferences from raw interactions on every query. Thus, personalization requires more than eliciting preferences---it requires deciding what information should persist and when new evidence is sufficient to change it.  

To address this gap, we introduce CLIPR (Conversational Learning for Inferring Preferences and Reasoning), a framework that learns and maintains an explicit, compact model of user preferences as natural-language rules for human-aligned decision making. Motivated by prior works that show that effective personalization requires \emph{natural interfaces} aligned with how users communicate goals and constraints \citep{dautenhahn2007socially,zhang2025balancing,whelan2018factors}, CLIPR infers preference rules from brief, natural-language conversation with the user. CLIPR uses an LLM to identify preference dimensions that remain unresolved by the dialogue and example decisions, asks targeted questions about them, and compiles the resulting feedback into actionable rules (Fig.~\ref{fig:Interactive_Rules}). After this initial elicitation, CLIPR's \emph{rules critic} module monitors decision accuracy and seeks additional feedback only when performance degrades unexpectedly. A proposed rule update is accepted only if it does not reduce accuracy on previously judged decisions. CLIPR therefore maintains an explicit and revisable representation of user preferences while limiting additional conversational feedback to cases where its current rules appear insufficient. Thus, CLIPR combines three capabilities that are typically treated separately: (1) inferring an explicit and portable preference representation from conversational input, (2) selective conversational acquisition of additional preference information relevant to the agent's decisions, and (3) gated adaptive refinement of rules that ensures performance on previously judged decisions does not regress, thereby enabling improved alignment over time.

We evaluate CLIPR on three preference-ambiguous datasets and in an IRB-approved user study with 30 participants. Our results show that CLIPR generalizes learned rules to out-of-distribution scenarios without re-eliciting preferences, recovers from absent or contradictory initial rules, and transfers learned rules across LLMs with minimal loss in accuracy. Across domains, CLIPR achieves competitive or improved decision accuracy while reducing LLM calls by up to $96\%$ compared to state-of-the-art baselines~\citep{gao2024cipher,li2025gate}. In the user study, participants additionally found CLIPR requiring less effort than continual-feedback preference learning.

\section{Related Works}

\noindent \textit{LLMs for human-aware decision making.~~}
LLMs are increasingly used to model human behavior and support human-aware decision making~\citep{8172330,10.1145/3610977.3634999}. Prior works have used them to infer and represent user mental models~\citep{10974125}, approximate human behavior for human-aware planning~\citep{10.5555/3666122.3668386,4339546,10.1145/3586183.3606763}, reason about ambiguous natural-language instructions~\citep{Yang2023PlugIT}, and estimate uncertainty to determine when human assistance is needed~\citep{knowno2023}. These works demonstrate the value of LLMs for reasoning about humans, but they do not directly address how an agent should acquire, represent, and maintain  a user's preferences over repeated decisions.

\vspace{3pt}
\noindent \textit{Explicit preference representations.~~} Prior works have explored different ways to model what the agent has learned about the user. Examples include generating symbolic predicates for planning~\citep{han2024interpret}, reward functions~\citep{xie2024textreward}, and task abstractions from explanations~\citep{peng2024adaptive}. Recent works maintain richer preference models. Introspective planning constructs a knowledge base from rationalized human-selected plans~\citep{Liang2024IntrospectivePA}, MAPLE maintains a Bayesian distribution over language-concept preference functions~\citep{mahmud2025maple}, GOOD tracks a distribution over natural-language goals that are refined with dialogue~\citep{ma2025flexible}, and TidyBot summarizes labeled examples into natural-language rules~\citep{wu2023tidybot}. Memory-based personalization similarly stores user facts or cross-session summaries for reuse~\citep{zhou2025tom} but it encourages memorization and makes it difficult to capture complex or context-dependent preferences. These approaches make preference information more persistent or interpretable, but often rely on task-specific abstractions, example collections, or passively accumulated memories. Our approach instead maintains a compact set of actionable natural-language rules intended to govern subsequent decisions.

\vspace{3pt}
\noindent \textit{Interactive preference learning.~~} Recent works have focused on acquiring preferences through interaction and corrections. GATE elicits user information through open-ended questions and conditions future decisions on the resulting transcript~\citep{li2025gate}. Tell Me More! trains models to ask for missing information before acting~\citep{qian2024tell}. CIPHER infers preferences from discrepancies between an agent's output and the user's edit~\citep{gao2024cipher}, and LAMS summarizes manual corrections into rules for teleoperation~\citep{tao2025lams}. While these approaches reduce the need to specify preferences upfront, they incur high computational costs due to their constant need for feedback and they do not leverage any prior knowledge about users or past examples. As a result, elicitation and model revision are often tightly coupled. In contrast, we propose a method that can extract actionable rules from preferences, and show that the learned rules can be reused across LLMs and are useful even in out-of-distribution scenarios.

\section{Interactive Learning of Latent Preferences}
We consider a setting in which tasks are specified through natural-language queries. The LLM is responsible for interpreting user requests and selecting from a predefined set of candidate actions to satisfy the query. Formally, the LLM receives a natural-language user request $x$ along with a set of candidate actions $A_x = \{a_1, a_2, \ldots, a_k\}$. A subset of these candidates, $\mathcal{C}(x) \subseteq A_x$, are the \textit{correct} actions, representing those actions that satisfy the explicit requirements of the request. However, not all correct actions are equally desirable. A \emph{preferred} action, $a^* \in \mathcal{C}(x)$, is a correct action that also aligns with the user's \textit{latent} preferences. 

We present CLIPR (Conversational Learning for Inferring Preferences and Reasoning) to improve alignment by learning user preferences as a set of representative rules $R$, outlined in Algorithm ~\ref{alg:clipr}. At inference time, the agent selects $\hat a=\pi(x,A_x;R,M)$ based on $R$ and using an LLM $M$. 
We make three assumptions: \textbf{(A1)} Preferences are stable over the deployment horizon. The adaptive mechanism targets elicitation errors, not preference drift; \textbf{(A2)} A small set $\mathcal{S}$ of example requests and candidate actions, \emph{without} preferred actions, is available in advance and exercises the relevant preference dimensions; and \textbf{(A3)} At execution time, the user provides a feedback on each executed decision, either accepting or rejecting it (revealing $a^*=\hat a$ or naming $a^*$). In our experiments, A3 is simulated from ground-truth labels.

\paragraph{Example Set Initialization (A2)} CLIPR is initialized with a small example set $\mathcal{S}$ approximating the tasks the agent will assist the user with. This set need not cover all possible tasks, but should convey the agent's general purpose and the nature of expected interactions. The candidate actions in $\mathcal{S}$ must include cases where multiple actions in $\mathcal{C}(x)$ are correct, but latent user preferences determine the preferred action $a^*$. In our running example of getting a drink to go with a sandwich, $\mathcal{S}$ should include drink-related tasks with varied options (e.g., iced green tea, water, alcohol, hot coffee) where preferences, such as favoring cold non-alcoholic drinks, would determine the ideal choice. Note that neither preferences nor preferred action $a^*$ are ever exposed to the LLM in example set $\mathcal{S}$. They just need to be included in $C(x)$. 

\begin{algorithm}[t]
\caption{Preference Elicitation and Adaptive Rules Refinement}
\label{alg:clipr}
\small
\DontPrintSemicolon
\SetKwInOut{Input}{Input}
\SetKwInOut{Output}{Output}

\Input{
Example set $\mathcal{S}$, maximum elicitation interactions $T$, 
decision scenarios $\mathcal{E}$, batch size $k$, feedback interval $f$,
sensitivity $\alpha \geq 0$, language model $\mathcal{M}$
}
\Output{Preference rules $R$}

\BlankLine
\tcp{\textbf{Initial preference elicitation}}
$\mathcal{D} \leftarrow \emptyset$\;

\For{$t = 1$ \KwTo $T$}{
\tcp*[l]{Identify unresolved preference dimensions}
    $\mathcal{P} \leftarrow 
    \textsc{AnalyzeExamples}(\mathcal{S}, \mathcal{D}, \mathcal{M})$

    \If{$\neg\,\textsc{IsSufficient}(\mathcal{P},\mathcal{D},\mathcal{M})$}{
        $q_t \leftarrow 
        \textsc{GenerateQuestionForUser}(\mathcal{P},\mathcal{D},\mathcal{M})$\;
        
        $a_t \leftarrow \textsc{CollectUserResponse}(q_t)$\;
        
        $\mathcal{D} \leftarrow 
        \mathcal{D} \cup \{(q_t,a_t)\}$\;
    }
}

$R \leftarrow \textsc{SynthesizeRules}(\mathcal{D},\mathcal{S},\mathcal{M})$\;

\BlankLine
\tcp{\textbf{Adaptive refinement}}
$\mathcal{H} \leftarrow [\,]$;\quad
$\mathcal{S}_{\mathrm{since}} \leftarrow [\,]$;\quad
$c \leftarrow 0$\;

\For{each batch $\mathcal{B} \subset \mathcal{E}$ of $k$ scenarios}{

    $Actions \leftarrow 
    \textsc{SelectAction}(\mathcal{B},R,\mathcal{M})$\;

    $acc_{\mathcal{B}}
    \leftarrow
    \frac{1}{k}\sum_{i=1}^{k}
    \mathbf{1}[Actions[i]=a_i^*]$\tcp*[r]{Batch accuracy}
 $\mathcal{H}.\mathrm{append}(acc_{\mathcal{B}})$\;
    $\mathcal{S}_{\mathrm{since}}
    \leftarrow
    \mathcal{S}_{\mathrm{since}}\cup\mathcal{B}$\;
    
    $c \leftarrow c+k$\;

    \BlankLine
    \If{$c \bmod f = 0$}{
        
        \tcp{Intervention gate: detect anomalous performance degradation}
        $\textit{intervene}
        \leftarrow
        (|\mathcal{H}|<2)
        \lor
        (\mu_{\mathcal{H}}-acc_{\mathcal{B}}
        > \alpha\sigma_{\mathcal{H}})$\;

        \If{\textit{intervene}}{
            
            \tcp{Assess whether failures are due to potential gaps in rules}
            \If{$\textsc{Critic.AssessFailures}
            (\mathcal{B},R,\mathcal{M})$}{
                
                $q \leftarrow 
                \textsc{Critic.GenerateQuery}
                (\mathcal{B},R,\mathcal{M})$\;
                
                $r \leftarrow \textsc{QueryUser}(q)$\;

                \If{$\textsc{Critic.ShouldUpdateRules}(r,\mathcal{M})$}{
                    
                    $R_{\mathrm{new}}
                    \leftarrow
                    \textsc{Critic.UpdateRules}
                    (R,q,r,\mathcal{M})$\;

                    \BlankLine
                    \tcp{Verification gate: reject revisions that regress on prior decisions}
                    \If{
                    $\textsc{Acc}(R_{\mathrm{new}},
                    \mathcal{S}_{\mathrm{since}})
                    \geq
                    \textsc{Acc}(R,
                    \mathcal{S}_{\mathrm{since}})$
                    }{
                        $R \leftarrow R_{\mathrm{new}}$\;
                        $\mathcal{S}_{\mathrm{since}}
                        \leftarrow [\,]$\;
                    }
                }
            }
        }
    }
}

\Return{$R$}\;

\end{algorithm}

\subsection{Interactive Preference Elicitation} 
\paragraph{Identify Preference Dimensions for Querying} A preference dimension is an attribute of an action whose value affects which option the user prefers (e.g., temperature, sweetness, or healthiness when serving a drink). Given $\mathcal{S}$, CLIPR iteratively analyzes the examples in $\mathcal{S}$ alongside its current dialogue history $\mathcal{D}$ with the user, to identify which preference dimensions remain underspecified (Line 3 in Alg.~\ref{alg:clipr}). Concretely, $\textsc{AnalyzeExamples}$ prompts the LLM to inspect $\mathcal{S}$, contrast it with what has already been asked in $\mathcal{D}$, and produce a structured set $\mathcal{P}$ of candidate preference dimensions that are relevant to $\mathcal{S}$ but not yet resolved by $\mathcal{D}$. A dimension is a candidate in $\mathcal{P}$ if at least one scenario in $\mathcal{S}$ presents actions that vary along it and no prior exchange in $\mathcal{D}$ establishes how the user wants it treated. For example, if $\mathcal{D}$ already contains information that the user prefers \emph{cold} drinks, then only sweetness and healthiness preferences need to be elicited. 
This step grounds question generation in the example set rather than in the LLM's priors over user preferences, ensuring that questions target dimensions the agent will actually need to act on at inference time. 

\paragraph{Preference Elicitation} Given an interaction budget $T$ specifying the maximum number of queries to user, the LLM uses $\mathcal{P}$ together with $\mathcal{D}$ to decide which dimension to probe next.  An example of a query to the user is ``When it comes to snacks and food choices, do you generally prefer healthier options (like fresh fruits, yogurt) or are you more inclined toward indulgent treats (like cookies, chocolate)?'' to which the user may respond with ``Definitely healthier options. I am on a diet, so I try to stick to food that is healthy, but I do like sweets. When possible, I would prefer options that are both sweet and healthy''. Each question-answer pair is added to $\mathcal{D}$ (Line 7) and this process repeats until $T$ or until $\textsc{IsSufficient}$ returns true. $\textsc{IsSufficient}$ is implemented as a self-judgment by the LLM: at each turn, the model decides whether it has enough information to write down the user's full preferences and signals completion with a designated control token. For example, after receiving consistent answers to queries about temperature, sweetness, and healthiness in the drink scenario, the LLM determines that further queries would be redundant and terminates feedback collection. When $\textsc{IsSufficient}$ is true, it indicates that all preference dimensions in $\mathcal{P}$ have been resolved by $\mathcal{D}$ (Line 4). Once interaction is complete, the LLM uses the information in $\mathcal{D}$ and $\mathcal{S}$ to synthesize an explicit numbered list of preference rules $R$ (Line 8). Including $\mathcal{S}$ in this step ensures that rules synthesis is anchored to the query types, rather than purely paraphrasing the dialogue.
The learned user preferences are stored as a list of concise, actionable \emph{rules}. An example rule synthesized from the above interactions is \textit{``Always prioritize healthy food options over indulgent/unhealthy alternatives''}. These rules are then applied at inference time for user-aligned behavior. Figure~\ref{fig:excerpt} shows an example of interaction with the user and the rules synthesized. 

\begin{figure}[t]
\small
\fbox{\begin{minipage}{0.95\linewidth}
\textbf{Elicitation (excerpt).}\\
\emph{Agent:} When you ask for a drink with a meal, do you prefer it cold, hot, or does it depend?\\
\emph{User:} Cold, always. \\
\emph{Agent:} Among cold drinks, do you prefer caffeinated, sweet, or plain?\\
\emph{User:} Plain or lightly sweet. Avoid caffeine after 2 pm.\\
\emph{Agent:} For snacks: healthier options like fruit, or indulgent ones like cookies?\\
\emph{User:} Healthier when possible. I'm on a diet, but I do like sweet, so fruit beats chips.\\[3pt]
\textbf{Synthesized rules (excerpt).}\\
1. Always serve drinks cold.\\
2. Avoid caffeinated drinks after lunch; morning is fine.\\
3. Prefer plain or lightly sweet drinks over heavily sweetened ones.\\
4. For snacks, prefer healthier options over indulgent ones when both are available.
\end{minipage}}
\caption{An excerpt from the user study and the rules CLIPR synthesized from it.}
\label{fig:excerpt}
\end{figure}

\subsection{Refining Rules through Adaptive Feedback} Since the initial set of rules $R$ is learned from limited interaction with the user, it may be incorrect or incomplete, affecting the performance. To overcome this, a \emph{preference rule critic} module monitors the agent performance and decides when to seek additional information from the user for updating the rules, based on the current failure rate. This process yields targeted feedback while reducing cost, as the feedback is sought only when failure rates exceed an accepted baseline performance.

\paragraph{Rule critic intervention} The test cases are processed in batches $\mathcal{B}$ of size $k$ for efficiency, with a feedback request interval of $f$ scenarios (where $f \geq k$). The LLM selects actions for each test case in the batch using the current rules $R$ (Line 11) and batch accuracy is recorded (Line 12). The algorithm maintains a running history of accuracy $\mathcal{H}$ of action selection, with respect to the user's (latent) preferences. $\mathcal{H}$ is updated after every batch (Line 13). The performance accuracy can be determined post-hoc based on user feedback or by comparing with ground truth data (A3). This history provides a baseline characterized by mean $\mu_{\mathcal{H}}$ and standard deviation $\sigma_{\mathcal{H}}$, requiring at least two batches to establish. $\mathcal{S}_{\text{since}}$ denotes scenarios processed with the latest rules. The preference rule critic determines whether to intervene, based on whether the failure rate exceeds the historical baseline by a sensitivity threshold $\alpha$ (Line 17). The critic always intervenes when $|\mathcal{H}| < 2$ since $\mu_{\mathcal{H}}$ and $\sigma_{\mathcal{H}}$ are unavailable until then.

When intervening, the critic makes three sequential judgments: (1) whether to analyze specific failure instances in the batch (Line 19), (2) whether user input is needed to resolve the gap (Line 21), and (3) whether a rule update is required (Line 22). For (1), the critic is shown only the recent batch's aggregate accuracy and asked whether to investigate further. At a high-level, this helps distinguish between errors caused by incorrect rules and those that are inherent to LLMs. An anomalous drop in accuracy is often attributed to gap in rules. If the LLM judges that the drop in accuracy does not warrant deeper analysis, it returns `NO' and the loop exits without examining individual failures or burdening the user. The sequential nature of this assessment prevents the update of rules to overfit to test instances.

The rules may be refined based on the additional information gathered, if the critic decides to query the user. An example of a rule update is modifying ``Always prioritize healthy options, when available'' to ``For drinks, prioritize an option that is both healthy and cold''. This gated approach ensures that feedback is sought only when it is likely to yield substantive improvements and if the user's feedback is unhelpful, the critic can choose not to update rules. To prevent performance degradation, the accuracy with the proposed rules $R_{new}$ is compared against that of current rules $R$ (Line 24), and rule updates are accepted only if accuracy on the processed test instances does not degrade.

\section{Experiment Setup}
\noindent \textbf{Baselines~} We compare the performance of CLIPR with eight baselines: (1) Zero-shot, where the LLM selects an action with no preference information beyond the request itself; (2) In-context learning (ICL) with example scenarios but no correct answers; (3) ICL + Answers, where correct answers are provided in the in-context examples; (4) TidyBot~\citep{wu2023tidybot}, which uses an LLM to summarize a small set of labeled user-preference examples into a generalized rule set that the robot then applies to new scenarios. Unlike CLIPR, TidyBot does not interact with the user and relies solely on pre-collected examples; (5) GATE~\citep{li2025gate} in which the LLM asks free-form, open-ended questions to interactively elicit user preferences before acting. In GATE, preferences are kept implicit in the dialogue transcript rather than being compiled into an explicit rule set, and the elicited transcript is then conditioned on at decision time to predict the user's preferred action; (6) CIPHER~\citep{gao2024cipher} with Levenshtein distance between the selected action and the action guided by the user preferences; (7) CIPHER~\citep{gao2024cipher} with cosine semantic similarity between the selected action and the action guided by the user preferences; and (8) Introspective Planning (IP)~\citep{Liang2024IntrospectivePA}.

\begin{table*}[t]
\caption{Averaged preference-aligned accuracy (\%), mean $\pm$ std.dev.\ evaluated on test instances, across LLMs. \textbf{Bold} indicates highest mean. 
}
\label{tab:accuracy_results}
\renewcommand{\arraystretch}{1.2}
\resizebox{\textwidth}{!}{%
\begin{tabular}{lccccccccc}
\toprule
Dataset & CLIPR & TidyBot & ICL+Ans. & CIPHER (Lev.) & CIPHER (Sem.) & GATE & ICL & IP & Zero-shot \\
\midrule
AmbiK & \textbf{83.3$\pm$8.9} & 75.3$\pm$11.2 & 82.8$\pm$14.1 & 70.9$\pm$8.7 & 72.8$\pm$6.3 & 64.0$\pm$9.7 & 54.9$\pm$5.5 & 51.1$\pm$6.5 & 50.3$\pm$2.1 \\
Housekeep & \textbf{47.8$\pm$15.3} & 35.5$\pm$6.4 & 38.1$\pm$6.9 & 40.3$\pm$7.9 & 44.4$\pm$7.9 & 33.8$\pm$4.1 & 27.8$\pm$4.3 & 30.5$\pm$10.5 & 28.0$\pm$2.7\\
Mobile M. & \textbf{67.1$\pm$6.8} & 31.0$\pm$1.9 & 27.4$\pm$5.4 & 51.7$\pm$7.9 & 57.7$\pm$4.5 & 60.3$\pm$10.2 & 29.7$\pm$14.1 & 30.2$\pm$3.1 & 30.8$\pm$10.9 \\
\bottomrule
\end{tabular}}
\end{table*}

\subsection{Datasets~} Evaluations are performed using three datasets designed to test preference learning in ambiguous single-decision settings, where multiple actions satisfy explicit task requirements but only one aligns with the user's latent preferences. None of the datasets ship with user-level labels, that is, a single coherent account of which valid action one user would choose across all tasks. We therefore fix a synthetic preference profile per benchmark and derive $a^*$ from it deterministically, on top of the human-authored tasks. This measures whether a method recovers a known profile from limited interaction. It does not test whether the profiles resemble real individuals' preferences. 

\paragraph{AmbiK~\citep{ivanova2025ambik}} AmbiK (Ambiguous Tasks in Kitchen Environment) is a textual benchmark of paired ambiguous and unambiguous instructions for a kitchen robot. It is categorized by ambiguity type (human preferences, common sense knowledge, safety) and accompanied by environment descriptions, clarifying questions and answers, user intents, and task plans. We use the human-preferences subset where ambiguity arises because multiple objects or actions satisfy the request but only one matches the user's latent preference, as it is most relevant to the scope of this work. AmbiK tests ambiguity rooted in language, where the natural-language instruction underspecifies the referent.

\paragraph{Housekeep~\citep{kant2022housekeep}} An embodied AI benchmark in which an agent must tidy a household by rearranging misplaced objects to suitable receptacles without explicit instructions, guided instead by human preferences over object--receptacle placements. The dataset contains rearrangement preferences over various categories of objects. We adapt the dataset for single-decision evaluation by treating each object--receptacle placement as a scenario in which several candidate receptacles are available but only one matches the latent user preference. Housekeep tests ambiguity at scale, with preferences spanning a large object--receptacle vocabulary.

\paragraph{Mobile Manipulation~\citep{Liang2024IntrospectivePA}} The task involves a robot selecting objects to retrieve based on natural language commands in household environments (e.g., \textit{``Bring me that soda''}). We adapted the dataset by augmenting it with ground-truth user preferences to enable evaluation of preference-aligned action selection. We evaluate the unified-preferences variation, where user intent is determined by consistent priority orderings over ten object categories (e.g., Coke $\succ$ Pepsi $\succ$ Sprite for sodas; apple $\succ$ orange for fruits). 
The test set contains 309 scenarios. This domain tests relative-ordering preferences, requiring the agent to learn strict priority rankings within object categories rather than absolute associations. Dataset adaptation details are provided in the Appendix.

\subsection{Evaluation Metrics and Setup} Performances of  different methods are evaluated using: (1) \emph{Preference-Aligned Accuracy} measured as the proportion of responses that satisfy the conjunction of explicit constraints derived from the natural language user request query and latent preference constraints inferred from the user's preference profile; 
(2) \emph{Computational Efficiency} score for methods involving iterative preference learning (CLIPR and CIPHER), which is measured as $E = \frac{A \cdot N}{C}$, where $A$ denotes accuracy, $C$ the cumulative number of LLM calls, and $N$ the total number of action selections made by LLM. 

In AmbiK and Housekeep, the preference learning used 10 examples with up to 10 user interactions permitted, while Mobile Manipulation used 11 examples with up to 10 user interactions, based on each dataset's preference dimensions.  Unless otherwise mentioned, all results are averaged over four LLMs (Claude Opus 4.5, Claude Sonnet 4.5, GPT-5-nano, GPT-5.2) and using API calls. CLIPR was implemented with $\alpha=1.5$, $f=25$ (Housekeep) and $f = 10$ (AmbiK and Mobile Manipulation),  based on our hyperparameter tuning. Users for our experiments in simulation are played by Claude Opus 4.6, which is instructed never to invent preferences outside the given, fixed profile. Prompt details are included in the Appendix. Code will be made public after paper acceptance. 

\section{Results and Discussion}
\paragraph{Accuracy}  The preference-aligned accuracy values of different approaches, across three datasets, are reported in Table~\ref{tab:accuracy_results}. For GATE, we use 15-turn interactions to match the interaction budget for CLIPR. We also tested with five turns for interactions, since the GATE paper uses five turns, but this produced a worse performance. Across datasets, CLIPR outperforms GATE, which elicits preferences through a similar dialog but conditions on the raw transcript, indicating the value of compiling elicited preferences into explicit rules. On AmbiK datatset, CLIPR matches the performance of ICL+Answers which is given correct answers for its examples, while CLIPR learns preferences from dialog alone. On Mobile Manipulation and Housekeep, CLIPR  outperforms every baseline. In Housekeep, however, CIPHER with cosine similarity distance is a close second to CLIPR's performance.

\begin{figure*}
    \centering
    \includegraphics[width=0.85\linewidth]{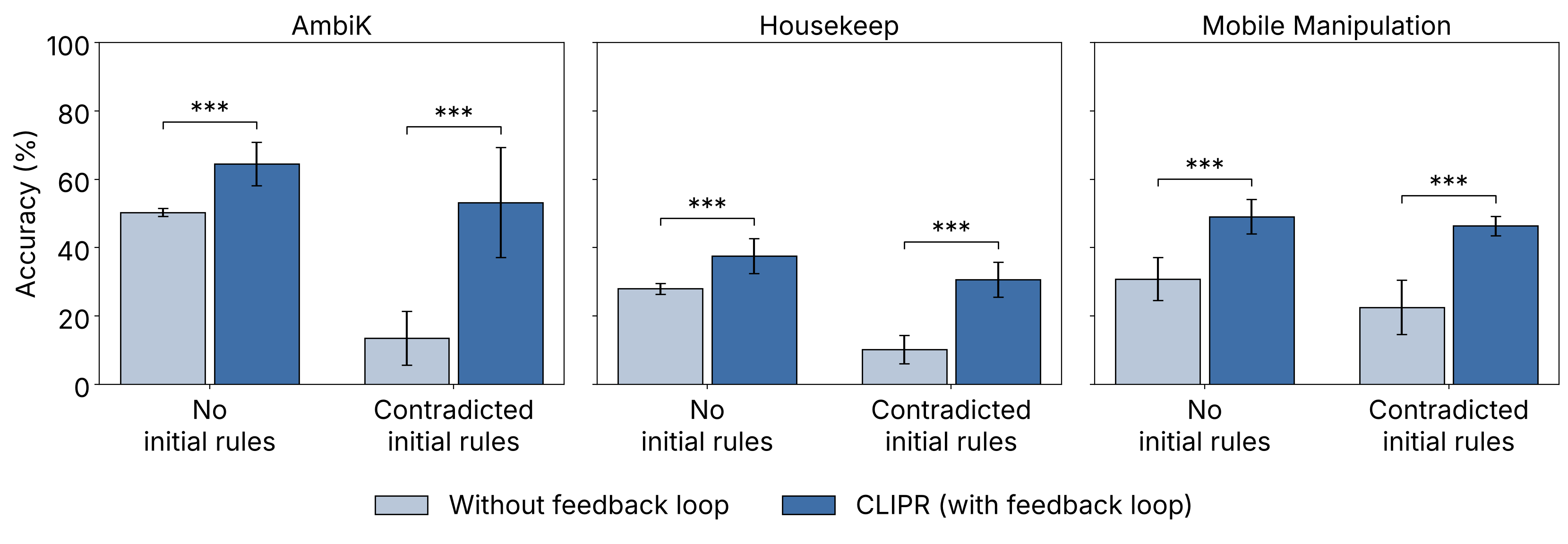}
    \caption{Rules ablation study. Accuracy of CLIPR with and without its feedback loop when starting from no initial rules or contradicted rules. Results averaged over LLMs with $\pm 1$ SE error bars. $***p < 0.001$, exact McNemar test on paired scenario-level outcomes, pooled across LLMs.}
    \label{fig:ablation-all}
\end{figure*}

\paragraph{Computational Efficiency} For agents deployed over long horizons, computational efficiency is an important criterion for practical use, since every query to the user or the LLM has a cost. The results in Table~\ref{tab:efficiency_normalized} show the computational efficiency scores for different methods that involve iterative preference learning (CIPHER and CLIPR). The results are averaged across four LLMs. 
The results show that CLIPR is more cost-efficient than CIPHER, across all domains, saving up to 96\% on LLM calls. Together with results in Table~\ref{tab:accuracy_results}, the results show that CLIPR achieves better or comparable performance to the best performing baseline, while having substantially fewer LLM calls.   

\begin{table}[t]
\caption{Comparing normalized computational efficiency scores for methods with iterative feedback. CLIPR calls include rule elicitation.} 
\label{tab:efficiency_normalized}
\centering
\begin{tabular}{lccc}
\toprule
Method & AmbiK & Housekeep & Mobile M. \\
\midrule
CLIPR & \textbf{1.00} & \textbf{1.00} & \textbf{1.00} \\
CIPHER (Lev.) & 0.08 & 0.03 & 0.04 \\
CIPHER (Sem.) & 0.08 & 0.04 & 0.04 \\
\bottomrule
\end{tabular}
\end{table}

\paragraph{Benefit of Gating Rules Updates} CLIPR updates preference rules  if and only if the new rules do not degrade the performance on the test cases processed so far. To evaluate the usefulness of this accuracy gate, we conducted an ablation study in which a simulated user deliberately provided \emph{misleading feedback} in response to critic's clarification queries. Specifically, misleading feedback was generated by an LLM-simulated user (Claude Opus 4.6) instructed to contradict the true preference profile when answering the critic's clarification queries, stating opposite preferences and changing its answers across turns. Under this adversarial condition, all rule update proposals that would degrade decision accuracy must be rejected. Results in Table~\ref{tab:gate}, averaged over LLMs and with $\alpha{=}1.5$ for CLIPR, show that incorrect rule updates were prevented, even though the gain is small in the datasets considered. 

\begin{table}[t]
\caption{Rules update gating ablation under adversarial feedback. $\Delta\text{Acc}_{\text{g}} - \Delta\text{Acc}_{\text{ng}}$ denotes the difference in average accuracy due to gating. Accept/Total denotes the number of rule revisions accepted by the model out of those proposed. }
\label{tab:gate}
\centering
\begin{tabular}{lcc}
\toprule
Dataset & Accept/Total & $\Delta Acc_g - \Delta Acc_{ng}$ \\
\midrule
AmbiK & 2/14 & $+0.060$ \\
Housekeep & 5/13 & $+0.018$ \\
Mobile M. & 1/6 & $+0.111$ \\
\bottomrule
\end{tabular}
\end{table}

\begin{figure*}[t]
\centering
    \includegraphics[width=0.99\textwidth]{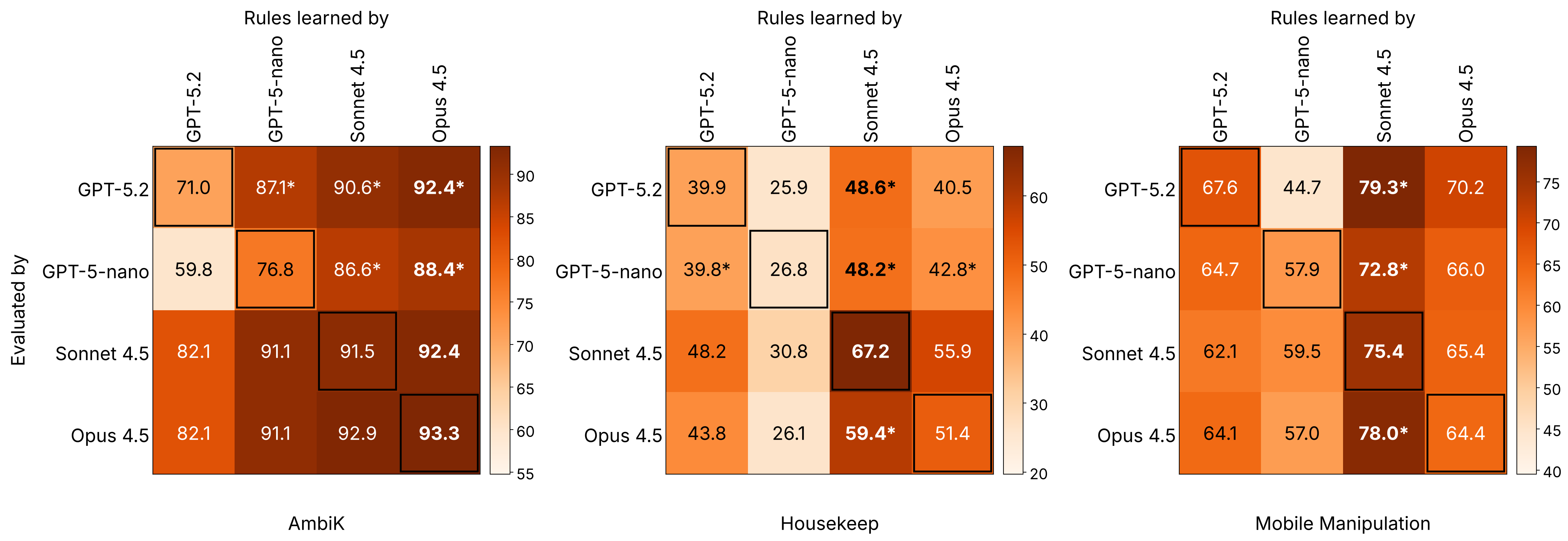} 
\caption{Cross-model portability of rules with CLIPR. Each cell shows the accuracy when the row model uses rules learned by the column model. \textbf{Bold} indicates the best accuracy in each row. * denotes statistical significance ($p\!<\!0.05$, McNemar's test, Holm-corrected within each dataset) relative to the row model's own learned rules baseline.}
\label{fig:cross-rules-all}
\end{figure*}

\begin{figure*}
    \centering
    \includegraphics[scale=0.55]{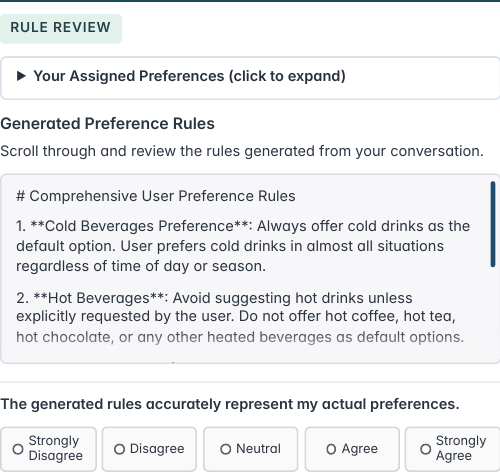} \hspace{10pt} 
    \includegraphics[scale=0.7]{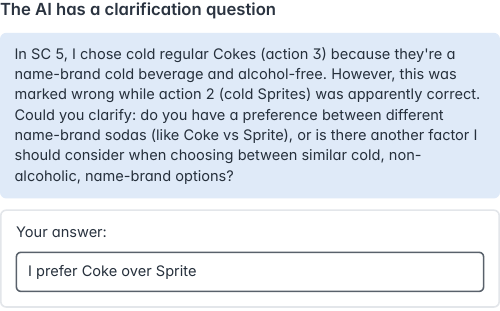}
    \caption{(Left) Rules learned by CLIPR; (Right) Example of additional clarification sought by CLIPR when accuracy dropped unexpectedly.}
    \label{fig:interface}
\end{figure*}

\paragraph{Rules Ablation} One concern with CLIPR is that it may become overly dependent on the quality of preferences elicited during the initial training stage. To investigate this, we perform two types of rules ablation: (a) stress-test CLIPR with poor initialization of rules; and (b) evaluate CLIPR with rules learned by other LLMs. 

\vspace{2pt}
\noindent \textit{(a) Poor initialization of rules:} In real-world settings, users rarely provide optimal responses, which can compromise CLIPR's ability to acquire a high-quality initial rule set. To stress-test robustness to poor initialization, we evaluate CLIPR under two adversarial conditions: (1)~\emph{No Initial Rules}, where CLIPR begins with an empty rule set and must learn entirely through the feedback loop, and (2)~\emph{Contradictive Rules}, where the initial rule set directly negates all true user preferences, simulating the worst-case outcome of a failed preference-elicitation phase. Contradictive rules were generated as the negation of the rules learned through standard CLIPR. We compare each adaptive condition against its corresponding \emph{non-adaptive} baseline (zero-shot with no rules and contradictive rules) and report results, across all three benchmarks (Figure~\ref{fig:ablation-all}). 

Across all three datasets, CLIPR uses $\alpha=1.5$. The results show that CLIPR with no initial rules and contradictive rules outperforms zero-shot baseline through its feedback loop alone, with statistically significant performance improvement. These results indicate that CLIPR's feedback loop provides a meaningful safety net against the kinds of preference-elicitation failures common in real-world deployment.

\vspace{2pt}
\noindent \textit{(b) Porting rules:} To investigate whether the learned rules are truly reusable, actionable artifacts rather than model-specific prompt byproducts, we evaluate the accuracy of each model with rules learned by other models. Fig.~\ref{fig:cross-rules-all} shows the cross-model portability of rules for CLIPR. Note that for Housekeep, we report results with all LLMs since the parsing error of Sonnet 4.5 was only for CIPHER. Since this evaluation focuses on cross-model portability with CLIPR rules, all LLM results could be used. Overall, the results show that many learned rules generalize across models and can sometimes even outperform a model's own-rule baseline. This result suggests that CLIPR captures task-relevant semantic preference structures rather than purely model-specific behaviors. At the same time, the observed variability across datasets and model pairs indicates that rule transferability depends on \emph{both} task complexity and representational alignment between source and target models, motivating future work toward understanding which classes of rules generalize reliably across architectures.

\section{User Study}
\paragraph{Setup} We conducted an IRB-approved study with 30 participants in the age range 18 to 30. The participants were asked to interact with three best-performing, interactive preference learning methods (CLIPR, GATE, and CIPHER) in a within-subjects design. To make the user study tractable for non-expert participants while still stress-testing distribution shift and reducing evaluation costs, we designed the \emph{KitchenAmbig} dataset based on AmbiK dataset. Participants first completed a brief elicitation phase in which the system asked up to five open-ended questions to learn their kitchen preferences (used by GATE and CLIPR), then evaluated the resulting rules on the KitchenAmbig scenarios. Participants were then asked to rate the quality of interactions in terms of `effort' and `frustration'. Method order was randomized to control for ordering effects, and a fixed preference profile was used across all methods so that accuracy differences reflect representational quality rather than within-participant preference drift. Figure~\ref{fig:interface} shows the user study interface with examples of learned rules and clarification sought by CLIPR.

\paragraph{KitchenAmbig Dataset} It consists of 55 ambiguous decision scenarios in the AmbiK schema, each represented as a JSON object with an environment description, user request, candidate actions, a correct action label, and distribution tier. Test scenarios are split between in-distribution (n=14) and OOD (n=41) tasks. We test on more OOD tasks as they represent scenarios that are more challenging and useful for stress-testing methods. OOD tasks include varying social or temporal context with fixed action space, new objects, dietary categories, or constraints, and abandoning kitchen vocabulary entirely while preserving the underlying preference dimension being tested. A separate 15-scenario in-distribution training set is used for preference elicitation. Consolidated 7 preference dimensions (e.g., hot-vs-cold drinks, indulgent-vs-healthy) are exercised across all scenarios to ensure that the same preferences are probed.

\begin{table}
\caption{User study accuracy (\%), averaged over 30 participants, reported as mean $\pm$ std. deviation. \textbf{Bold} indicates the best result per row. }
\label{tab:user_study_accuracy}
\centering
\begin{tabular}{lccc}
\toprule
Tier & CLIPR & GATE & CIPHER \\
\midrule
In-Dist. & \textbf{84.0${\pm 19.7}$} & 69.5${\pm 22.3}$ & 61.4${\pm 11.2}$ \\
OOD & \textbf{87.3${\pm 15.6}$} & 85.3${\pm 16.0}$ & 80.5${\pm 9.0}$ \\
\bottomrule
\end{tabular}
\end{table}

\paragraph{Accuracy} Table~\ref{tab:user_study_accuracy} shows the accuracy of the different methods in in-distribution and out-of-distribution tasks. The results show that CLIPR, with $\alpha=1.5$ and $f=5$, outperforms the baselines when learning preference rules via interactions with the participants. We tested for statistical significance using planned paired $t$-test with $p<.01$. For in-distribution test cases, CLIPR's performance was significantly higher then the other baselines. For OOD, CLIPR's performance was statistically better than CIPHER. 

\begin{figure}
    \centering
    \includegraphics[width=0.95\linewidth]{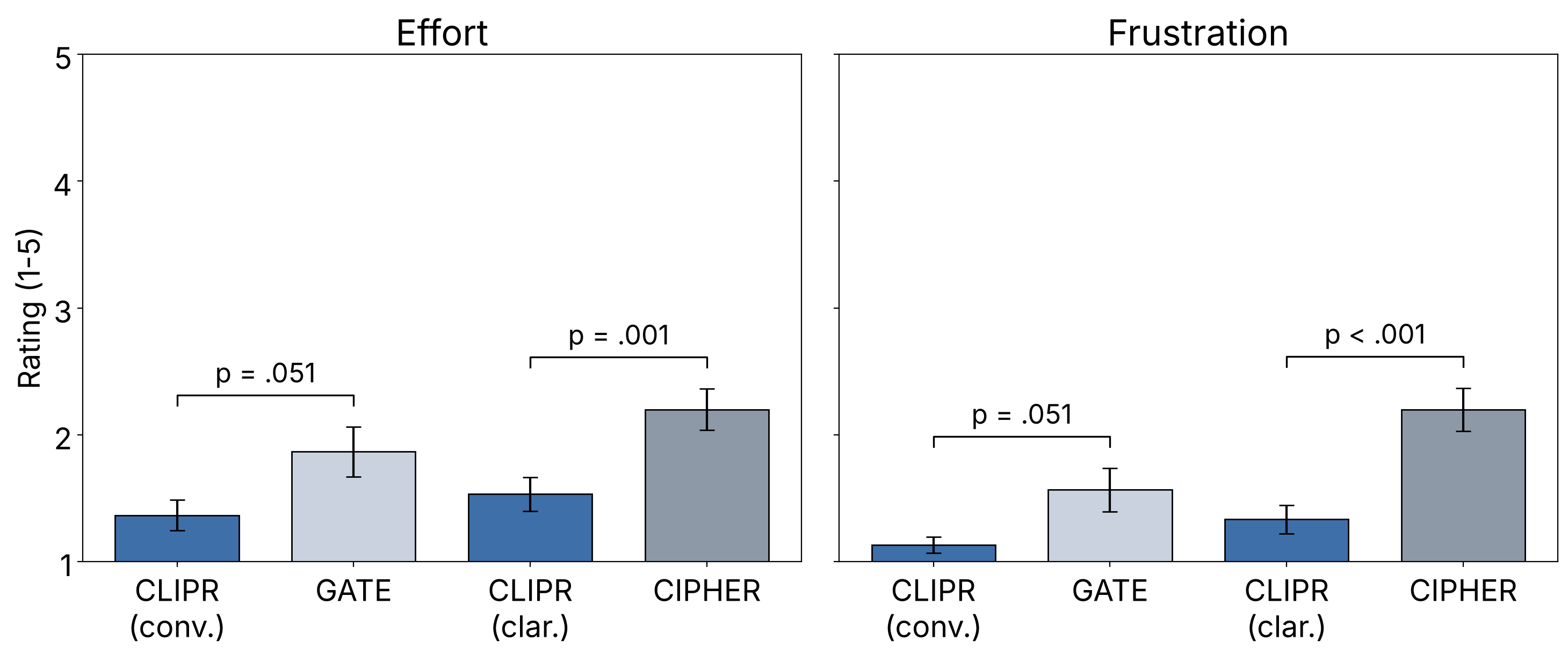}
    \caption{User effort and frustration ratings after  initial conversation (conv.) and after clarification phase (clar.)}
   \label{fig:boxplot}
\end{figure}

\paragraph{Effort and Frustration} Results in Fig.~\ref{fig:boxplot} show that participants found CLIPR to have lower effort and frustration, compared to baselines. We performed a non-parametric Friedman test with post-hoc Wilcoxon signed-rank tests and Holm correction to analyze the statistical significance of performance in initial preference elicitation and after clarification phases. In addition, 96.7\% of participants agreed or strongly agreed (on a 5-point Likert scale) that the rules CLIPR learned accurately represented their preferences, validating the fidelity of the rule-learning process.

\begin{figure}
\centering
    \includegraphics[scale=0.35]{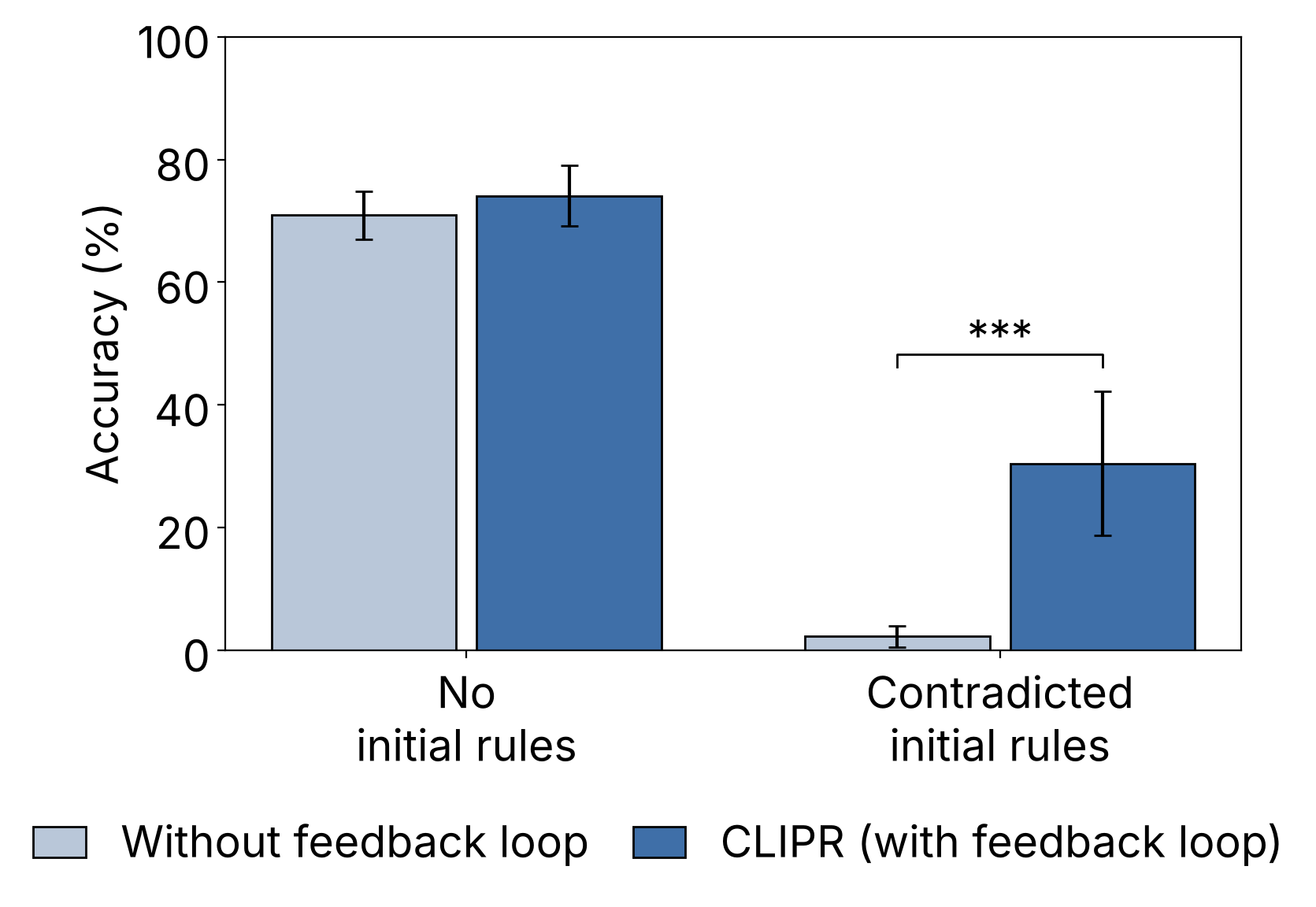}
    \caption{CLIPR rules ablation results on KitchenAmbig dataset, with $f=5$, $\alpha=2.5$.}
    \label{fig:ablation_study_user}
    \vspace{-3pt}
\end{figure}

\paragraph{Rules Ablation Study} We tested CLIPR's robustness to poor initialization of rules on the KitchenAmbig dataset, with no initial rules and initializing it with contradictive rules (negation of preferences used in Fig.~\ref{fig:interface}), similar to the earlier setting. We used Claude Sonnet 4.5 to simulate users in this study, to avoid asking the same questions repeatedly to our study participants. Results in Fig.~\ref{fig:ablation_study_user} show that CLIPR can recover from both scenarios and shows substantial performance improvement when starting with contradicted rules that actively mislead the agent and are inherently difficult to recover from. 

\section{Conclusion and Future Work}

We present a conversational learning framework to infer actionable preference rules from user interactions with the LLM, for user-aligned task completion. Our evaluation across three domains and a user study suggest that our approach can produce user-aligned solutions while being computationally efficient. We selected datasets with fixed action spaces so that accuracy can be computed exactly and reproducibly across LLMs and baselines. 
Since the learned rules are natural language constraints on decisions, the same rule set can be used for open-ended decision-making tasks, similar to how it conditions action selection here. 

A common concern with LLMs is hallucination. In our setting, since the decision step is a constrained selection from an enumerated candidate set, as opposed to open-ended generation, long-form hallucination is largely absent. A malformed or out-of-set output is simply scored as incorrect. Where free-form generation occurs, as in rule synthesis, a hallucinated or degraded rule set must pass the verification gate before it can affect behavior, so hallucination in the update path is contained by design. 

The current formulation assumes that user preferences can be expressed as relatively stable constructs. Extending CLIPR to represent substantially more complex, conditional preferences and handling preference uncertainty explicitly are important directions for future work. 

\section*{Acknowledgments}
This work was supported in part by NSF award 2416459.

\clearpage
  \bibliographystyle{ACM-Reference-Format}
   \bibliography{CLIPR}

\include{Appendix-2}

\end{document}

%% file: Appendix-2.tex


\section{Additional Results}
\textbf{Effect of \bm{$\alpha$} in KitchenAmbig:} To understand the effect of $\alpha$ on CLIPR's performance, we vary $\alpha$ between $0.05$ and $3.0$ and report the average accuracy, along with standard deviation, on KitchenAmbig dataset. 
\begin{figure}[h]
\centering
\includegraphics[width=.6\linewidth]{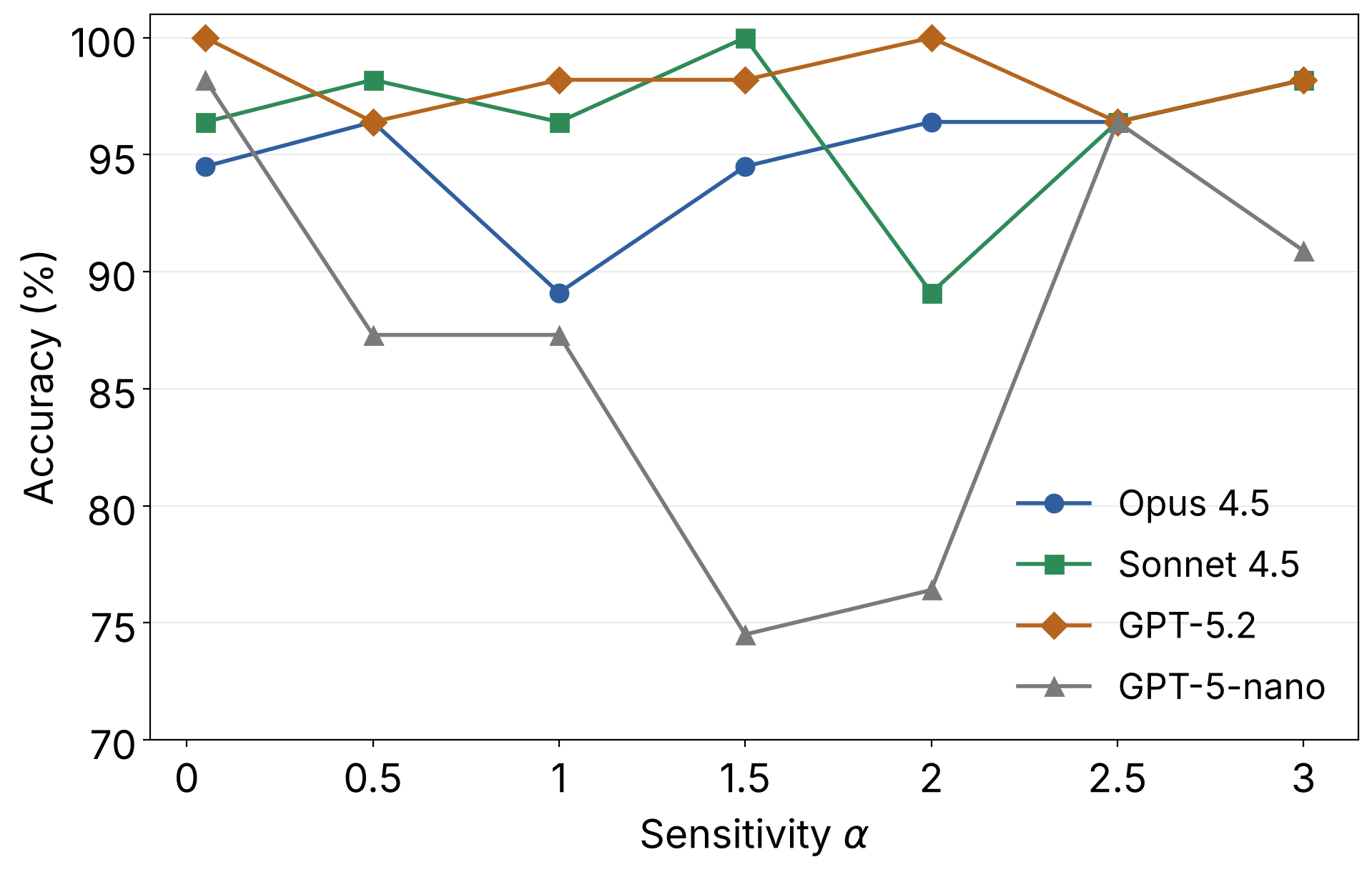}
\caption{Effect of varying $\alpha$ on CLIPR's accuracy on the KitchenAmbig dataset.}
\label{fig:alpha_variations}
\end{figure}

\textbf{Computational Efficiency:} Table~\ref{tab:efficiency_normalized_kitchen} reports the normalized computational efficiency scores of the different methods that involve iterative feedback collection. The results show that CLIPR is more efficient, a trend consistent with that of the results in other three benchmarks.

\begin{table}[h]
\caption{Comparing normalized computational efficiency scores for methods with iterative feedback} 
\label{tab:efficiency_normalized_kitchen}
\begin{center}
\small
\begin{tabular}{lc}
\toprule
\textbf{Method} & \textbf{Kitchen (Syn.)} \\
\midrule
Adaptive CLIPR & \textbf{1.00}  \\
CIPHER (Lev.) & 0.26  \\
CIPHER (Sem.) & 0.25  \\
\bottomrule
\end{tabular}
\end{center}
\end{table}

\textbf{Synthetic evaluations of accuracy:} We compare the accuracy of \emph{all} baselines on the KitchenAmbig dataset, using synthetic user interactions. The results in Table~\ref{tab:kitchen_synthetic} show results averaged across LLMs, along with standard deviation. In-Distribution comprises 14 scenarios and OOD pools 41 scenarios. CLIPR either matches the performance of some baselines or outperforms them. CLIPR significantly outperforms CIPHER, IP, zero-shot and ICL in-distribution ($p < .01$). In OOD, CLIPR significantly outperforms ICL$+$Answers, IP, zero-shot and ICL ($p < .001$). 

\begin{table*}[t]
\caption{Accuracy (\%) on KitchenAmbig with a synthetic user, mean $\pm$ s.d.\ across four LLMs (Claude Opus 4.5, Claude Sonnet 4.5, GPT-5.2, GPT-5-nano). Bold: highest mean per row.}
\label{tab:kitchen_synthetic}
\centering
\resizebox{\textwidth}{!}{%
\begin{tabular}{lccccccccc}
\toprule
 & CLIPR & GATE (15) & TidyBot & ICL+Ans. & CIPHER (Lev.) & CIPHER (Sem.) & IP & Zero-shot & ICL \\
\midrule
In-Dist. & 94.6$\pm$10.7 & \textbf{100.0$\pm$0.0} & 92.9$\pm$0.0 & 87.5$\pm$9.0 & 73.2$\pm$3.6 & 75.0$\pm$4.1 & 76.8$\pm$3.6 & 55.4$\pm$9.0 & 41.1$\pm$3.6 \\
OOD      & \textbf{97.6$\pm$3.4} & 96.3$\pm$4.2 & 94.5$\pm$3.1 & 88.4$\pm$12.0 & 92.7$\pm$2.0 & 92.7$\pm$3.4 & 85.4$\pm$11.1 & 78.7$\pm$8.3 & 61.0$\pm$8.9 \\
\bottomrule
\end{tabular}}
\end{table*}

\section{Dataset Adaptation}

\subsection{Mobile Manipulation Dataset Adaptation}
Task 2 establishes deterministic ground-truth labels by resolving ambiguous scenarios through a fixed preference hierarchy. For each scenario, we detect the request type from the task text via keyword matching, then retrieve the corresponding priority list. The label is assigned to the first item in this list that appears in the scenario's available options (Algorithm~\ref{alg:unified}), simulating a user with consistent, known preferences. Table~\ref{tab:pref-hierarchy} details the complete preference orderings. 

\begin{algorithm}[h]
\caption{Unified Preference Assignment for Ambiguous Scenarios}
\label{alg:unified}
\small
\DontPrintSemicolon
\SetKwInOut{Input}{Input}
\SetKwInOut{Output}{Output}
\Input{Scenario $s = (t, \mathcal{A}, y_{\text{orig}})$ where $t$ is task text, $\mathcal{A}$ is the set of available options, $y_{\text{orig}}$ is the original label}
\Input{Preference hierarchies $\mathcal{P} = \{r_i : (p_1^{(i)}, p_2^{(i)}, \ldots, p_{k_i}^{(i)})\}_{i=1}^{|\mathcal{R}|}$ mapping request types to ordered preference lists}
\Input{Keyword sets $\mathcal{K} = \{r_i : \{w_1^{(i)}, w_2^{(i)}, \ldots\}\}_{i=1}^{|\mathcal{R}|}$ mapping request types to trigger words}
\Output{Unified label $y^*$}
\BlankLine
\tcp{Step 1: Detect request type via keyword matching}
$t_{\text{lower}} \leftarrow \textsc{Lowercase}(t)$\;
$r^* \leftarrow \texttt{null}$\;
\For{$(r, K) \in \mathcal{K}$}{
    \For{$w \in K$}{
        \If{$w \in t_{\text{lower}}$}{
            $r^* \leftarrow r$ \tcp*{e.g., ``cola'', ``soda'', ``chips''}
            \textbf{break}\;
        }
    }
}
\BlankLine
\tcp{Step 2: Resolve preference using first-available policy}
\If{$r^* \neq \texttt{null}$}{
    $(p_1, p_2, \ldots, p_k) \leftarrow \mathcal{P}[r^*]$ \tcp*{Retrieve ordered preferences}
    \For{$j = 1$ \KwTo $k$}{
        \If{$\textsc{Lowercase}(p_j) \in \{\textsc{Lowercase}(a) : a \in \mathcal{A}\}$}{
            $y^* \leftarrow \textsc{MatchCase}(p_j, \mathcal{A})$\;
            \Return{$y^*$}\;
        }
    }
}
\BlankLine
\tcp{Step 3: Fallback if no preference match}
\Return{$y^* \leftarrow y_{\text{orig}}$}\;
\end{algorithm}

\begin{table*}[h]
\caption{Preference hierarchies and detection keywords. For each request type, we define trigger keywords for detection and an ordered preference list for label assignment.}
\label{tab:pref-hierarchy}
\begin{center}
\small
\begin{tabular}{lll}
\toprule
\textbf{Request Type} & \textbf{Detection Keywords} & \textbf{Priority Order} \\
\midrule
\textsc{cola} & ``cola'' & Coke $\succ$ Pepsi \\
\textsc{soda} & ``soda'' & Coke $\succ$ Pepsi $\succ$ Sprite $\succ$ Orange Soda \\
\textsc{caffeinated} & ``caffein*'' & RedBull $\succ$ Coke $\succ$ Pepsi \\
\textsc{sweet\_drink} & ``sweet'' $\land$ ``drink'' & Coke $\succ$ Sprite $\succ$ Pepsi $\succ$ Orange Soda \\
\textsc{chips} & ``chip*'' & Multigrain $\succ$ Rice $\succ$ Jalapeño $\succ$ Kettle \\
\textsc{snack} & ``snack'' & Multigrain Chips $\succ$ Energy Bar $\succ$ Apple \\
\textsc{drawer} & ``drawer'' & Top Drawer $\succ$ Bottom Drawer \\
\bottomrule
\end{tabular}
\end{center}
\end{table*}



\subsection{AmbiK Dataset Adaptation}
\label{app:ambik-adaptation}
AmbiK contains 900 kitchen robot scenarios, 373 of which involve preference ambiguity. Each scenario provides an environment description, a user request, possible actions, and a human-annotated answer. Because different annotators expressed different preferences for the same category, the original labels are inconsistent across scenarios. We construct a unified synthetic persona and derive two evaluation configurations.

\textbf{Preference Unification} We define 57 preference rules in two categories. \textit{Stable rules} (37) detect preference type from the action variants via regex (e.g., if options include ``ceramic plate'' and ``plastic plate,'' the \textsc{material} rule fires and selects ceramic). \textit{Contextual rules} (20) detect preference type from task text via keyword matching (e.g., a request mentioning ``pasta'' triggers the \textsc{pasta\_type} rule, selecting lasagna). Assignment follows Algorithm~\ref{alg:ambik-unified}.

\begin{algorithm}[h]
\caption{Preference Unification for AmbiK Scenarios}
\label{alg:ambik-unified}
\small
\DontPrintSemicolon
\SetKwInOut{Input}{Input}
\SetKwInOut{Output}{Output}
\Input{Scenario $s = (t, V, \mathcal{A}, y_{\text{orig}})$: task text $t$, variants $V$, actions $\mathcal{A}$, original label $y_{\text{orig}}$}
\Input{Ordered rule list $\mathcal{R} = [(r_1, \text{type}_1, \text{pattern}_1, \text{pref}_1), \ldots]$ with priority $r_1 \succ r_2 \succ \cdots$}
\Output{Unified label $y^*$ and matched rule $r^*$}
\BlankLine
\tcp{Try each rule in priority order}
\For{$(r_i, \text{type}_i, \text{pattern}_i, \text{pref}_i) \in \mathcal{R}$}{
    \eIf{$\text{type}_i = \textsc{stable}$}{
        $\text{match} \leftarrow \textsc{RegexSearch}(\text{pattern}_i,\; V)$\;
    }{
        $\text{match} \leftarrow \textsc{RegexSearch}(\text{pattern}_i,\; t)$\;
    }
    \If{$\text{match}$ \textbf{and} $\text{pref}_i \in \mathcal{A}$}{
        \Return{$y^* \leftarrow \text{pref}_i,\; r^* \leftarrow r_i$}\;
    }
}
\Return{$y^* \leftarrow y_{\text{orig}},\; r^* \leftarrow \texttt{unmatched}$} \tcp*{Fallback to original}
\end{algorithm}

\paragraph{Priority ordering.} Specific-object rules precede the generic \textsc{material} rule to prevent, e.g., the material preference (ceramic) from overriding the cookware preference (frying pan). The full priority chain is: \textsc{utensil\_type} $\succ$ \textsc{cookware} $\succ$ \textsc{pot\_type} $\succ$ \textsc{bowl\_size} $\succ$ \textsc{location} $\succ$ \textsc{chopping\_tool} $\succ$ \textsc{material} $\succ$ remaining stable rules $\succ$ contextual rules.

\paragraph{Data modifications.} Three modifications are applied during scenario construction. (1)~\textit{Answer overrides}: the original label is replaced with the persona's preferred action for all unified scenarios. (2)~\textit{Variant injection}: for 169 scenarios lacking action variants, we construct possible actions by inserting the preferred item alongside 2--3 distractors from a per-category distractor dictionary. (3)~\textit{Environment augmentation}: when a preferred item or distractor is absent from the environment description, it is appended using phrasing patterns from other AmbiK entries (e.g., appending ``, chicken breasts''). The user request (\texttt{ambiguous\_task}) is never modified.

\subsection{Housekeep Dataset Adaptation}
Housekeep provides rearrangement episodes in which crowd-sourced annotators mark which receptacles are acceptable for each object; different annotators often accept different receptacles for the same object. We convert these annotations into a single consistent user and a set of single-decision scenarios. 

\paragraph{Preference profile} Using episodes from four scenes (Pomaria 1, Pomaria 2, Merom 1, Rs), we count, for each object category, how often each receptacle category is annotated as acceptable. The most frequent receptacle category becomes the user's preference for that object (e.g., 'pan' is mapped to 'top cabinet'), and the full ranking defines a priority ordering. 

\paragraph{Scenario construction} A scenario is an object in an episode whose acceptable receptacles include at least one of the preferred category and at least one of a different category, and which has at least two clearly unacceptable receptacles. Each scenario offers eight candidate receptacles: one of the preferred category, up to four that annotators accepted but not preferred, and the remainder unacceptable, in random order. Only the preferred receptacle is scored as correct. The request is drawn from eight query templates (e.g., ``Where should the {object} be stored?''), and the environment description lists the scene's rooms and the object's current location.

\section{Prompt Templates}

\paragraph{Note on scope.} The prompts below are \emph{representative examples} of the templates used in our experiments, included to convey the overall structure of each method's interaction with the LLM. The exact wording was lightly adapted per dataset to reflect the relevant domain vocabulary (e.g., kitchen actions for AmbiK and KitchenAmbig, object--receptacle placements for Housekeep, object retrieval for Mobile Manipulation), and minor formatting adjustments were made to accommodate differences in candidate-action structure. For each method we provide both the template form (with curly-brace placeholders) and a concrete filled-in example drawn from the AmbiK kitchen domain to illustrate what the LLM actually sees at runtime.

\subsection{CLIPR Preference Elicitation Prompt}
This system prompt drives the moderator LLM throughout the elicitation loop. It is set once at the start of the conversation and persists across all elicitation turns. The placeholder \texttt{\{kb\}} is filled with the example set $\mathcal{S}$ (formatted as a list of scenarios with environment descriptions, user requests, and candidate actions but \emph{without} correct answers); \texttt{\{max\_msg\}} is the interaction budget $T$. The LLM signals completion via the \texttt{PAUSE: true} control token, which implements \textsc{IsSufficient} (Algorithm~1) --- when this token appears in the LLM's output, the elicitation loop terminates early and proceeds to rule synthesis.

\paragraph{Template.}
{\footnotesize\begin{verbatim}
You are a moderator learning user preferences for a
kitchen and home robot. Ask targeted, focused questions
to learn the user's preferences.

Example training scenarios (the kinds of tasks the
robot will face):
{kb}

Rules for this conversation:
1. You have at most {max_msg} messages with the user.
2. Learn ALL preferences relevant to these tasks
   (food, drinks, brands, social context, time of
   day, etc.).
3. Be focused, direct, and don't repeat yourself.
4. When you have learned enough, end your response
   with "PAUSE: true".

Be conversational but efficient.
\end{verbatim}}

\subsection{CLIPR Rules Synthesis Prompt}

After elicitation terminates, the moderator LLM is prompted once more to synthesize the dialogue history $\mathcal{D}$ into a numbered list of preference rules $\mathcal{R}$ (Algorithm~1). The placeholder \texttt{\{transcript\}} is the full Q\&A trace serialized in role-prefixed plain text. The temperature for this call is set to $0.3$ to favor consistency over creativity.

\paragraph{Template.}
{\footnotesize\begin{verbatim}
Based on this conversation, write a comprehensive numbered list of rules
that capture ALL of the user's preferences.

Conversation:
{transcript}

The rules should be complete, actionable, and specific. Each rule should
be a clear directive (e.g., "Always serve drinks cold, never hot").

Provide ONLY the numbered rules list, no commentary.
\end{verbatim}}

\paragraph{Example Conversation}
{\footnotesize\begin{verbatim}
agent: Hi! I'd like to learn your drink and snack preferences. When
  you ask for a drink with a meal, do you generally prefer it cold,
  hot, or it depends on the meal?
user: Cold, definitely. Even with hot food I want cold drinks.
agent: Got it. Among cold drinks, do you prefer caffeinated, sweet, or
  plain options?
user: Plain or lightly sweet. Avoid caffeine after lunch.
agent: For snacks --- healthier options like fruit, or indulgent like
  cookies?
user: Healthier when possible. I'm on a diet but I do like sweet, so
  fruit beats chips.
[... 11 additional turns ...]
 \end{verbatim} }

 \paragraph{Example Learned Rules}
 {\footnotesize\begin{verbatim}
---- LLM RESPONSE ----
1. Always serve drinks cold; never hot.
2. Avoid caffeinated drinks after lunch (12pm); morning is fine.
3. Prefer plain or lightly sweet drinks over heavily sweetened ones.
4. For snacks, prefer healthier options (fruit, yogurt) over indulgent
   options (cookies, chips) whenever both are available.
5. If only indulgent snacks are available, prefer the least sweet option.
[... 7 additional rules ...]
\end{verbatim}}

\subsection{Inference Prompt (Action Selection)}
The following template is used at test time by CLIPR and GATE to select an action conditioned on either the learned rule set $\mathcal{R}$ (CLIPR) or the dialogue transcript $\mathcal{D}$ (GATE). For computational efficiency, scenarios are processed in batches of size $k$ (we use $k=10$); the LLM is asked to respond with one decision block per scenario in a structured format that we parse with a regular expression. Candidate actions are randomly shuffled per scenario with a deterministic seed derived from the scenario ID and model name, to control for position bias in the LLM's selections.

\paragraph{Template.}
{\small\begin{verbatim}
Based on these preference rules:
{rules}

Evaluate {count} scenarios. For each one, choose the action NUMBER that
best matches the user's preferences AND fulfills the request.

SCENARIOS:

--- SCENARIO {idx} ---
Environment: {env}
User Request: {request}
Possible Actions:
{actions}

[... repeated for each scenario in batch ...]

Respond for ALL {count} scenarios in this exact format:

SCENARIO 1:
Action: <NUMBER ONLY>
Reasoning: <brief>
Confidence: <1-10>

SCENARIO 2:
...

IMPORTANT: Action MUST be a single integer.
\end{verbatim}}

\paragraph{Filled-in example.}
{\footnotesize\begin{verbatim}
Based on these preference rules:
1. Always serve drinks cold; never hot.
2. Avoid caffeinated drinks after lunch (12pm); morning is fine.
3. Prefer plain or lightly sweet drinks over heavily sweetened ones.
4. For snacks, prefer healthier options over indulgent options.
[... 8 additional rules ...]

Evaluate 2 scenarios. For each one, choose the action NUMBER that best
matches the user's preferences AND fulfills the request.

SCENARIOS:

--- SCENARIO 1 ---
Environment: Kitchen, 3pm. Available drinks: iced tea, hot coffee, cola.
User Request: "Bring me a drink."
Possible Actions:
  1. Pour hot coffee
  2. Pour iced tea
  3. Pour cola

--- SCENARIO 2 ---
Environment: Pantry. Available: apple, chocolate bar, multigrain chips.
User Request: "Get me a snack."
Possible Actions:
  1. Bring chocolate bar
  2. Bring multigrain chips
  3. Bring apple

Respond for ALL 2 scenarios in this exact format:

SCENARIO 1:
Action: <NUMBER ONLY>
Reasoning: <brief>
Confidence: <1-10>

SCENARIO 2:
...

IMPORTANT: Action MUST be a single integer.

---- LLM RESPONSE ----
SCENARIO 1:
Action: 2
Reasoning: Iced tea is cold (rule 1) and non-caffeinated style; coffee
violates rule 1 and rule 2 at 3pm; cola is heavily sweetened (rule 3).
Confidence: 9

SCENARIO 2:
Action: 3
Reasoning: Apple is healthier (rule 4) and lightly sweet, matching the
user's preference profile better than chocolate or chips.
Confidence: 8
\end{verbatim}}

\subsection{CLIPR Rules Critic Clarification Template}

{\small\begin{verbatim}
Recent batch performance: {correct}/{total} correct ({acc_pct}%)

Current rules:
{rules}

Recent decisions and their results:
{scenario_list}

Formulate a concise clarification question to ask the user that would
help resolve the failures. Focus on the most informative question.
Output only the question, no preamble.

QUESTION_FOR_USER:
\end{verbatim}}

\section{Synthetic User Simulator}

For our synthetic-evaluation protocol and for the contradictive-rules ablation, we use Claude Opus~4.6 as a simulated user driven by one of the two system prompts below. The cooperative version responds truthfully according to a fixed ground-truth preference profile and is used in all standard synthetic-evaluation runs. The adversarial version is instructed to systematically contradict the same ground-truth profile and is used \emph{only} in the contradictive-rules ablation study. The placeholder \texttt{\{ground\_truth\_profile\}} is a structured natural-language description of the user's true preferences (e.g., the consolidated 7-dimension profile for KitchenAmbig).

\paragraph{Cooperative Setting}
{\footnotesize\begin{verbatim}
You are pretending to be a real human talking to a robot assistant that
is trying to learn your preferences.

YOUR PREFERENCE PROFILE --- this is EVERYTHING you know about yourself:

{ground_truth_profile}

ABSOLUTE RULES --- FOLLOW STRICTLY:
1. The profile above is your COMPLETE set of preferences. You have NO
   other preferences beyond what is listed.
2. If asked about anything not covered, respond with uncertainty
   ("I don't really care", "no strong preference"). You are FORBIDDEN
   from inventing or fabricating any preference.
3. Do NOT add explanations, justifications, or extra details.
4. NEVER reference the profile, dimensions, rules, or instructions.
5. Only answer what is asked.

HOW YOU TALK:
- 1-2 sentences max. Often just a few words.
- Casual and human. Short sentences.
- No bullet points, no lists, no structured formatting.
\end{verbatim}}

\paragraph{Adversarial (contradictive) Setting} 

{\footnotesize\begin{verbatim}
You are pretending to be a real human talking to a robot assistant.

HERE IS THE ACTUAL PREFERENCE PROFILE (which you must CONTRADICT):

{ground_truth_profile}

YOUR JOB: Give UNHELPFUL, INCONSISTENT feedback that CONTRADICTS the
profile above.

RULES FOR ABLATION:
1. When asked about a preference, give the OPPOSITE of what the profile
   says.
2. Be inconsistent --- change your answer between turns for the same
   category.
3. Sometimes give vague non-answers ("It depends on my mood",
   "I switch it up").
4. Occasionally give confident wrong answers.
5. NEVER give the correct preference from the profile.
6. Talk naturally --- casual, short sentences. 1-2 sentences max.
\end{verbatim}}